\documentclass[runningheads]{llncs}
\usepackage{lipsum} 
\usepackage{xcolor}
\SetLipsumParListSurrounders{\begingroup\color{gray}}{\endgroup}
\usepackage{graphicx}
\usepackage{amsmath}
\usepackage{multirow}
\usepackage{booktabs}
\usepackage{amssymb}
\usepackage{amssymb}    
\usepackage{subcaption}
\usepackage{pifont} 
\usepackage{floatrow}
\usepackage{hyperref}
\hypersetup{
  colorlinks, linkcolor=red
}
\newfloatcommand{capbtabbox}{table}[][\FBwidth]
\usepackage{blindtext}

\begin{document}
\title{Classroom Slide Narration System}
\titlerunning{Classroom Slide Narration System}
\author{Jobin K.V. \and Ajoy Mondal \and C. V. Jawahar}
\authorrunning{Jobin et al.}
\institute{IIIT-Hyderabad, India \\
\email{\{jobin.kv@research., ajoy.mondal@, and jawahar@}\}iiit.ac.in \\
\url{http://cvit.iiit.ac.in/research/projects/cvit-projects/csns}}
\maketitle              

\begin{abstract}
Slide presentations are an effective and efficient tool used by the teaching community for classroom communication. However, this teaching model can be challenging for the blind and visually impaired ({\sc vi}) students. The \textsc{vi} student required a personal human assistance for understand the presented slide. This shortcoming motivates us to design a Classroom Slide Narration System ({\sc csns}) that generates audio descriptions corresponding to the slide content. This problem poses as an image-to-markup language generation task. The initial step is to extract logical regions such as title, text, equation, figure, and table from the slide image. In the classroom slide images, the logical regions are distributed based on the location of the image. To utilize the location of the logical regions for slide image segmentation, we propose the architecture, Classroom Slide Segmentation Network ({\sc cssn}). The unique attributes of this architecture differs from most other semantic segmentation networks. Publicly available benchmark datasets such as WiSe and SPaSe are used to validate the performance of our segmentation architecture. We obtained $9.54\%$ segmentation accuracy improvement in WiSe dataset.  We extract content (information) from the slide using four well-established modules such as optical character recognition (\textsc{ocr}), figure classification, equation description, and table structure recognizer. With this information, we build a Classroom Slide Narration System ({\sc csns}) to help {\sc vi} students understand the slide content. The users have given better feedback on the quality output of the proposed {\sc csns} in comparison to existing systems like Facebook’s Automatic Alt-Text ({\sc aat}) and Tesseract.  

\end{abstract}

\section{Introduction} \label{sec:intro}

Slide presentations play an essential role in classroom teaching, technical lectures, massive open online courses ({\sc mooc}), and many formal meetings. Presentations are an efficient and effective tool for the students to understand the subject better. However, the visually impaired ({\sc vi}) students do not benefit from this type of delivery due to their limitations in reading the slides. Human assistants could help {\sc vi} students to access and interpret the slide's content. However, it is impossible and impractical for the {\sc vi} students to have a human reader every time; because of the cost and the limited availability of trained personnel. From this perspective, this would be a primary issue in modern classrooms.

\begin{figure}[!ht]
\includegraphics[width=12cm,trim={0.4cm  
18.3cm 
7.6cm 
3cm 
},clip]{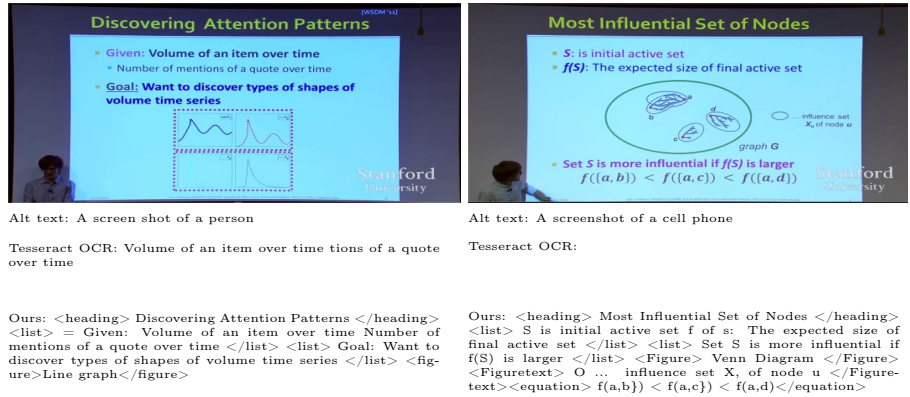}
\caption{Illustrate qualitative comparison between outputs obtained by the proposed {\sc csns} and the existing assistive systems --- Automatic Alt-Text ({\sc aat})~\cite{wu2017automatic} and Tesseract {\sc ocr}~\cite{smith2007overview}. The proposed system generates a markup text, where each logical content is tagged with its appropriate logical function. The audio file generated using this markup text gives a clear picture of the classroom slide's content to the blind or \textsc{vi} students. \label{fig:teaser}}
\end{figure}

The automatic extraction of logical regions from these slide images is the initial step for understanding the slide's content. This can be defined as, segmentation of semantically similar regions, closely related to natural image semantic segmentation in computer vision~\cite{long2015fully,chen2017deeplab,zhao2017pyramid}. The goal is to assign logical labels like title, paragraph, list, equation, table, and figure to each pixel in the slide images. Due to the large variability in theme (i.e., the layout), style, and slide content, extracting meaningful regions becomes a challenging task. The text extraction from these regions is beneficial for the \textsc{vi} student only if the text is tagged with its logical function. The left image of Fig.~\ref{fig:teaser}, the plain text ``\textit{Discovering Attention Patterns}'' is more meaningful to the \textsc{vi} student only if he knows the text is the heading of the slide. Similarly, detecting figures, equations, and table regions is also essential for understanding the slides. Our goal is to develop a system that meaningfully describes the classroom slides with a markup text, as shown in Fig.~\ref{fig:teaser}.

\vspace{-1cm}

The classroom slides have a complex design. The exact text could appear in various labels such as ``heading'', ``paragraph'', and ``list''. Hence, only the visual features learned by the existing semantic segmentation networks (e.g., {\sc fcn}~\cite{long2015fully}, Deeplab~\cite{chen2017deeplab}, and {\sc pspn}et~\cite{zhao2017pyramid}) are not sufficient to distinguish various logical regions. Limited works~\cite{haurilet2019spase,haurilet2019wise} on classroom slide segmentation utilize $(x,\ y)$ coordinate values to enhance the quality of segmentation results. However, all these works~\cite{haurilet2019spase,haurilet2019wise} do not utilize location information efficiently for segmentation. In contrast, Choi \emph{et al.}~\cite{Choi_2020_CVPR} utilize the position encoding~\cite{vaswani2017attention} to impose the height information in {\sc han}et for semantic segmentation of urban-scene images. In the case of slide images, the height and width information of each logical region are equally crucial for segmenting regions accurately. In this work, we propose a Classroom Slide Segmentation Network, called {\sc cssn}et to segment slide images accurately. The proposed network consists of (i) Attention Module, which utilizes the location encoding of the logical region (height and width of a region), and (ii) Atrous Spatial Pyramid Pooling ({\sc aspp}) module, which extracts multi-scale contextual features. The experiments on publicly available benchmark datasets WiSe~\cite{haurilet2019wise} and SpaSe~\cite{haurilet2019spase} establish the effectiveness of the proposed {\sc lean}et over state-of-the-art techniques --- Haurilet \emph{et al.}~\cite{haurilet2019wise}, Haurilet \emph{et al.}~\cite{haurilet2019spase}, {\sc han}et~\cite{Choi_2020_CVPR}, {\sc dan}et~\cite{fu2019dual}, and {\sc dran}et~\cite{fu2020scene}.     

The proposed Classroom Slide Narration System ({\sc csns}) creates audio content for helping {\sc vi} students to read and understand the slide’s content. The audio created corresponds to the segmented region's content in reading order. We use four existing recognizers ---{\sc ocr}~\cite{smith2007overview}, equation descriptor~\cite{mondal2019textual}, table recognizer~\cite{sachin2020table}, and figure classifier~\cite{jobin2019docfigure} to recognize content from the segmented regions. We evaluate the effectiveness of the developed system on the WiSe dataset~\cite{haurilet2019wise} and compare the user experiences with the existing assistive systems \textsc{aat}~\cite{wu2017automatic} and Tesseract {\sc ocr}~\cite{smith2007overview}. The proposed system attains a better user experience than the existing assistive systems. 

The contributions of this work are as follows:
\begin{itemize}
\item Proposes a {\sc cssn}et to efficiently segment the classroom slide images by utilizing the location encoding.
\item Presents a {\sc csns} as an use case of slide segmentation task to narrate slide's content in reading order. It helps the {\sc vi} students to read and understand the slide's content. 
\item Perform experiments on publicly available benchmark datasets WiSe and SPaSe establish the effectiveness of the proposed {\sc cssn}et and {\sc csns} over existing slide segmentation and assistive techniques.  
\end{itemize}

\section{Related Work}

\paragraph{{\bf Visual Assistance System:}}

According to the World Health Organization (\textsc{who}), at least $2.2$ billion people have a vision impairment or blindness worldwide. The computer vision community has undertaken significant efforts to deploy automated assistive systems for such people. Some of the unique assistive systems based on computer vision are --- (i) v{\sc oic}e\footnote{https://www.seeingwithsound.com}
~\cite{auvray2007learning} vision technology is a sensory substitution system for blind people which gives visual experience through live camera views by image-to-sound renderings, (ii) Electro-Neural Vision System ({\sc envs})~\cite{meers2004vision} provides a virtual perception of the three-dimensional profile and color of the surroundings through 
pulses, (iii) a wearable travel aid for environment perception and navigation~\cite{bai2019wearable}, (iv) a smartphone-based image captioning for the visually and hearing impaired people~\cite{makav2019smartphone}. (v) Currency Recognition using Mobile Phones~\cite{singh2014currency} (vi) Facebook’s Automatic Alt-Text (\textsc{aat})~\cite{wu2017automatic} system is one of the most widely used assistive features, which automatically recognizes a set of $97$ objects present in Facebook photos, (vii) FingerReader~\cite{shilkrot2015fingerreader} is a wearable text reading device to assist blind people and dyslexic readers. All these existing assistive systems fail to properly narrate the classroom slides for a {\sc vi} student due to various logical regions present in the slides. 

\paragraph{{\bf Document Image Segmentation:}}

Textual regions and graphics regions dominate document images. Graphics include natural images, various types of graphs, plots, and tables. Historically, the document image segmentation task is considered as dividing a document into various regions. Existing techniques~\cite{amin2001page,ha1995recursive} are either bottom-up or top-down. In the case of bottom-up~\cite{amin2001page}, individual words are first detected based on hand-crafted features. Therefore, detected words are grouped to form lines and paragraphs. Segmenting textual regions (like paragraphs) into lines and words helps to find semantic regions present in the document in the case of top-down~\cite{ha1995recursive}. These methods mostly use feature engineering and do not necessitate any training data while applying clustering or thresholding.

With the recent advancement in deep convolutional neural networks ({\sc cnn}s), several neural-based models~\cite{chen2015page,vo2016dense,renton2017handwritten,wick2018fully,yang2017learning} have been proposed to segment document images. Chen {\em et al.}~\cite{chen2015page} considered a convolutional auto-encoder to learn features from cropped document image patches. Thereafter, those features are used to train an {\sc svm} classifier~\cite{cortes1995support} to segment document patches.
Vo {\em et al.}~\cite{vo2016dense} and Renton {\em et al.}~\cite{renton2017handwritten} proposed lines detection algorithm in the handwritten document using {\sc fcn}. In~\cite{yang2017learning}, Yang {\em et al.} proposed an end-to-end, multi-modal, fully convolutional network for extracting semantic structures from document images. Inspired by document image segmentation and natural image segmentation using deep {\sc cnn}s, Haurilet \emph{et al.}~\cite{haurilet2019wise,haurilet2019spase} explore DeepLab~\cite{chen2017deeplab} architecture on classroom slide segmentation task. More recently, \textsc{dan}et~\cite{fu2019dual}, \textsc{han}et~\cite{Choi_2020_CVPR}, and \textsc{dran}et~\cite{fu2020scene} capture long-range dependency to improve performance by extending the self-attention mechanism. The proposed {\sc lean}et is strongly influenced by \textsc{dan}et~\cite{fu2019dual}, \textsc{han}et~\cite{Choi_2020_CVPR}, and \textsc{dran}et~\cite{fu2020scene}.    

\section{Classroom Slide Narration System} \label{system}

We develop a {\sc csns} as a use case of slide segmentation to help {\sc vi} students to read and understand slide's content without human assistance. The proposed {\sc csns} consists of three modules --- (i) \textbf{Slide Segmentation Module:} to identify logical regions present in the slide, (ii) \textbf{Information Extraction Module:} to extract meaningful information from each segmented region, and (iii) \textbf{Audio Creation Module:} to create an audio narration of extracted information in proper order. We discuss each of these modules in detail.

\begin{figure*}[!ht]
\includegraphics[width=12cm,trim={
0.25cm  
 1cm 
 0.51cm 
 1cm 
 },clip]{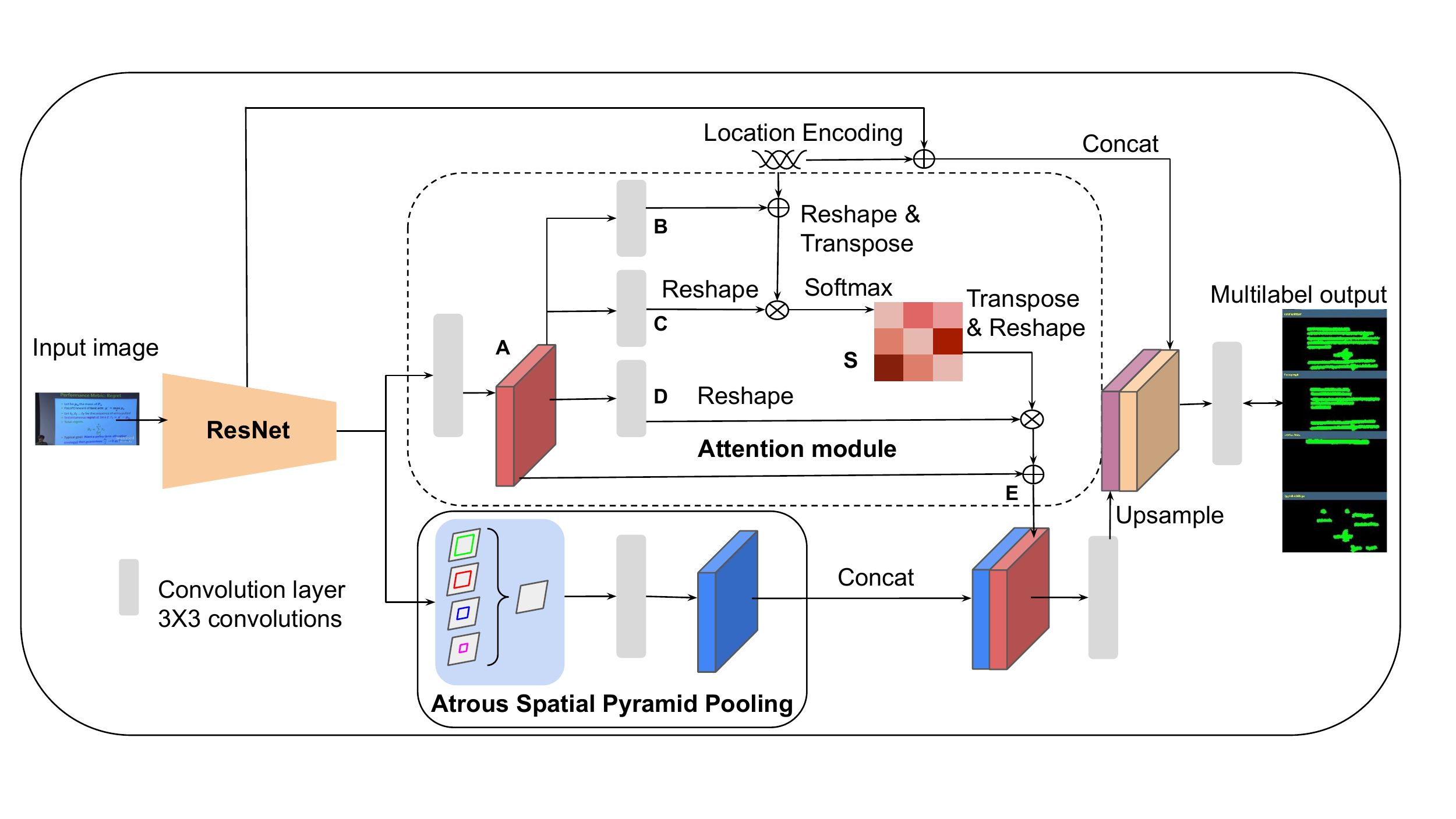}
\caption{Illustrate the architecture of the proposed {\sc cssn}et for classroom slide segmentation. The network consists of three modules --- (i) attention module (upper dotted region), (ii) multi-scale feature extraction module (lower  region), (iii) feature concatenation module. Here, $\oplus$ and $\otimes$ represent the element-wise summation and multiplication of features, respectively.\label{ProposedArch}}
\end{figure*}

\subsection{Slide Segmentation Module}\label{slide_segmentation}

{\sc cssn}et is the segmentation network utilized in this work to segment classroom slide images accurately. The {\sc cssn}et aims to learn more discriminating features for semantic associations among pixels through the location encoding attention mechanism. We propose a location encoded attention module that utilizes the location encoding of logical regions of slide image. To enhance the segmentation accuracy, we concatenate the attention module output feature with the multi-scale features through an Atrous Spatial Pyramid Pooling ({\sc aspp}) layer~\cite{chen2017deeplab} as shown in the Fig~\ref{ProposedArch}. Finally, a simple decoder network predicts the high-resolution, multi-label, dense segmentation output. 

\subsubsection{Attention Module:}

The previous attempts~\cite{chen2017deeplab,zhao2017pyramid} suggest that the local features generated by traditional Fully Convolutional Networks ({\sc fcn}s) could not capture rich context, leading to erroneous object detection. While Choi \emph{et al.}~\cite{Choi_2020_CVPR} discussed that urban-scene images have common structural priors depending on a spatial position. The authors also discussed height-wise contextual information to estimate channels weights during pixel-level classification for urban-scene segmentation. Fu \emph{et al.}~\cite{fu2019dual} also discussed the self-attention mechanism to capture the spatial dependencies between any two positions of the feature maps. Inspired by the works~\cite{Choi_2020_CVPR,fu2019dual}, we introduce an attention module that utilizes location encoding to enhance the local feature's contextual information and relative position. The attention module selectively encodes a broader range of contextual and location information into each pixel according to semantic-correlated relations, thus enhancing the discrimination power for dense predictions. 

As illustrated in Fig.~\ref{ProposedArch} (inside the dotted box),  we use a convolution layer to obtain features of reduced dimension $\mathbf{A} \in$ $\mathbb{R}^{C \times H \times W}$. We feed $\mathbf{A}$ into a convolution layer to generate two new feature maps $\mathbf{B}$ and $\mathbf{C}$, respectively. $\{\mathbf{B}, \mathbf{C}\} \in \mathbb{R}^{C \times H \times W}$. We add the location encoding $\mathbf{L}$ (refer Section~\ref{loc_enc}) to the $\mathbf{B}$ and obtain $\mathbf{L}+\mathbf{B}$. Then we reshape it to $\mathbb{R}^{C \times N}$, where $N=H \times W$ is the number of pixels. We perform a matrix multiplication between the transpose of $\mathbf{C}$ and $\mathbf{B}+\mathbf{L}$, and apply a soft-max layer to obtain spatial attention map $\mathbf{S} \in \mathbb{R}^{N \times N}$:
\begin{equation}
s_{j i}=\frac{\exp \left((B_{i}+L_{i}) \cdot C_{j}\right)}{\sum_{i=1}^{N} \exp \left((B_{i}+L_i) \cdot C_{j}\right)},
\end{equation}
where $s_{j i}$ measures the impact of $i^{th}$ position on $j^{t h}$ position. A more similar feature representation of two positions contributes greater correlation between them. Meanwhile, we feed feature $\mathbf{A}$ into a convolution layer to generate a new feature map $\mathbf{D} \in \mathbb{R}^{C \times H \times W}$ and reshape it to $\mathbb{R}^{C \times N}$. Then we perform a matrix multiplication between $\mathbf{D}$ and transpose of $\mathbf{S}$ and reshape the result to $\mathbb{R}^{C \times H \times W}$. Finally, we multiply it by a scale parameter $\alpha$ and perform an element-wise sum with the features $\mathbf{A}$ to obtain the final output $\mathbf{E} \in \mathbb{R}^{C \times H \times W}$ using:
\begin{equation}
E_{j}=\alpha \sum_{i=1}^{N}\left(s_{j i} D_{i}\right)+(1-\alpha)\times A_{j},
\end{equation}
where $\alpha$ is initialized as 0 and learned gradually.

\paragraph{\textbf{Location Encoding}}\label{loc_enc}

\begin{figure}[!htp]
\includegraphics[width=12cm,trim={
 0cm  
 9cm 
 0cm 
 2cm 
 },clip]{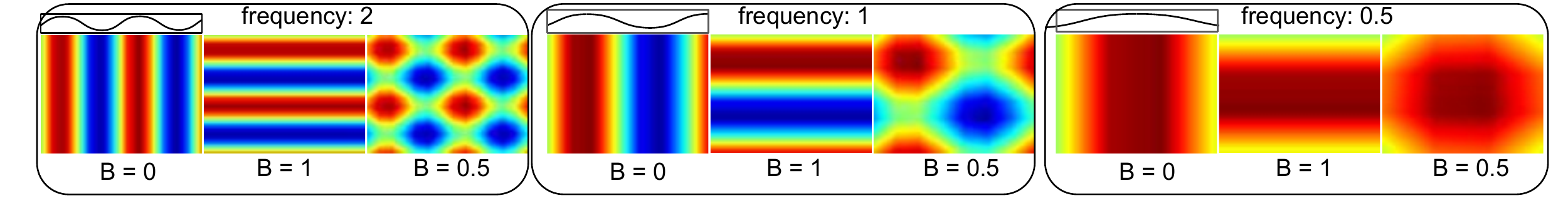}
\caption{Presents visualization of various location encoding with frequency $2$, $1$, and $0.5$ with $\beta$ values $0$, $0.5$, and $1$.} \label{locationEnc}
\end{figure}

We follow strategy proposed by~\cite{vaswani2017attention,Choi_2020_CVPR} for location encoding and use sinusoidal position encoding~\cite{vaswani2017attention} as $\mathbf{L}$ matrix. In our approach, we give importance to both height and width location encoding. The dimension of the positional encoding is the same as the channel dimension $C$ of the intermediate feature map that combines with $\mathbf{L}$. The location encoding is defined as
$$
\begin{aligned}
P E_{v}(v, 2 i) &=\sin \left(v / 100^{2 i / C}\right);
\ \ P E_{v}(v, 2 i+1)&=\cos \left(v / 100^{2 i / C}\right), \\
P E_{h}(h, 2 i) &=\sin \left(h / 100^{2 i / C}\right); 
\ \ P E_{h}(h, 2 i+1)&=\cos \left(h / 100^{2 i / C}\right),
\end{aligned}
$$
where $v$ and $h$ denote the vertical and horizontal location index in the image ranging from $0$ to ${H/r}-1$ and $0$ to ${W/r}-1$, and $i$ is the channel location index ranging from $0$ to ${C}-1$. The $r$ is called the reduction ratio, which determines the frequency of $\sin$ or $\cos$ wave. Here, we apply $\sin$ and $\cos$ location encoding for intermediate layers of the features. The new representation $\mathbf{L}$ incorporating location encoding is formulated as
$$
\mathbf{L}= \beta P E_{h} \oplus (1-\beta)P E_{v},
$$
where $\oplus$ is an element-wise sum, $\beta $ is initialized with $0$ and learned gradually. Adding both the horizontal and vertical position encoding enables the attention module to learn the location information in both directions. The $\beta$ value is learned based on the importance of width and height locations. Fig.~\ref{locationEnc} shows the location encoding value of a layer in $\mathbf{L}$ matrix. Finally, we add random jitters to the location encoding $PE_v$ and $PE_h$ for the better generalization of location encoding. The height and width of the location encoding matrix $\mathbf{L}$ are adjusted to the dimension of the feature vector.

\subsection{Information Extraction Module}\label{ie_block}

We coarsely group the classroom slide image into four logical regions --- text, figure, equation, and table and use existing recognizers to recognize the content of the regions.

\paragraph{\textbf{OCR:}}
We use well-known Optical Character Recognition ({\sc ocr}), Tesseract~\cite{smith2007overview} to recognize the content of the text regions present in slide images.    

\paragraph{\textbf{Figure Classification:}} Figures mainly different types of plots, sketches, block diagrams,maps, and others. appear in slides. We use DocFigure~\cite{jobin2019docfigure} to find the category of each figure in slide images.  

\paragraph{\textbf{Equation Description:}}
The equation frequently occurs in the classroom slides. We follow {\sc med} model~\cite{mondal2019textual} for describing equation in natural language to interpret the logical meaning of equation for {\sc vi} students.

\paragraph{\textbf{Table Structure Recognition:}}
Recognition of the table present in the slide includes detecting cells, finding an association between cells, and recognizing cell's content. We use TabStruct-Net~\cite{sachin2020table} to recognize the physical structure of the table and then Tesseract~\cite{smith2007overview} to recognize content. 

The information from the four modules is combined and saved as a {\sc json} file. These processes are done on the server and then sent to the {\sc json} file to the mobile app, and the {\sc vi} student understands the slide's content by interacting with the app.   

\subsection{Audio Creation Module}

Based on a study that empirically investigates the difficulties experienced by {\sc vi} internet users~\cite{murphy2008empirical}, a mobile application is developed for {\sc vi} students to read and understand the classroom slide's content. For this, we place a camera in the classroom connected to the server for capturing slide images. The {\sc vi} students use this mobile app to connect to the server through a WiFi connection. Using the mobile application (\textbf{CAPTURE} button), the student can request the server to get information on the current slide. The server captures the current slide image through the camera, extracts the information as {\sc json} file, and sends it back to the mobile application. Fig.~\ref{fig_mobile_application}(a) shows a screenshot of the developed mobile app with one example. The app has two modes --- (i) interactive and (ii) non-interactive.  In the interactive mode, the user can navigate various logical regions of slides by themselves. The app shows the captured slide image with all extracted information. When the user touches one particular area, it automatically plays the audio sound using android {\sc tts} library of that region's content along with the category label. In the non-interactive mode, the app automatically plays the audio corresponding to the slide's content. In this case, the screen contains the \textbf{READ ALL} button; when the user touches this particular area, the app automatically plays the audio sound of the full slide's content. These features are demonstrated in the demo video\footnote{\url{https://youtu.be/PnPYrA8ykF0}}

\section{Experiments}

\subsection{Experimental Setup}

We use the residual network (e.g., ResNet-101) pre-trained on ImageNet~\cite{krizhevsky2012imagenet} as the backbone network for all the experiments. We replace a single $7\times 7$ convolution with three of $3\times 3$ convolutions in the first layer of ResNet-101. We use We also adopt an auxiliary cross-entropy loss in the intermediate feature map. Since for the multi-label segmentation task we need to predict multiple classes per pixel, we replace the softmax output layer in the previous deep models with a sigmoid activation function and train these models using binary cross entropy loss. We train the network using \textsc{sgd} optimizer with an initial learning rate of $2e-2$ and momentum of $0.9$. We set the weight decays of $5e-4$ and $1e-4$ for backbone networks and segmentation networks, respectively. The learning rate schedule follows the polynomial learning rate policy~\cite{liu2015parsenet}. The initial learning rate is multiplied by $(1-\frac{iteration}{max\ iteration})^{0.9}$. We use typical data augmentations such as random scaling in the range of $[0.5,2]$, Gaussian blur, color jittering, and random cropping to avoid over-fitting. We set crop size and batch size of $540\times 720$ and $2$, respectively. We use mean Intersection over Union (mIoU)~\cite{chen2017deeplab} and Pixel Accuracy ({\sc pa})~\cite{haurilet2019wise,haurilet2019spase} for evaluation purpose.

\subsection{Dataset}

\paragraph{\textbf{WiSe:}}

The WiSe~\cite{haurilet2019wise} dataset consists of $1300$ slide images. It is divided into a training set, validation set, and test set consisting of $900$, $100$, and $300$ slide images annotated with $25$ overlapping region categories. 

\paragraph{\textbf{SPaSe:}} 

The SpaSe~\cite{haurilet2019spase} dataset contains dense, pixel-wise annotations of $25$ classes for $2000$ slide images~\cite{haurilet2019spase}. It is split into training, validation, and test sets containing $1400$, $100$, and $500$ images, respectively. 

\subsection{Ablation Study}

In the ablation study, we use the ResNet-101 backbone with an output stride of $16$ and evaluate the WiSe dataset. First, we examine the best feature vector for adding location encoding in the \textsc{lean}et architecture. Based on Fig.~\ref{ProposedArch} of {\sc cssn}et architecture, the location encoding can be added to any of the feature vectors including, $\mathbf{A}$, $\mathbf{B}$, $\mathbf{C}$, and $\mathbf{D}$. We empirically find the best feature for adding location encoding. In addition to it, we also conduct experiments by changing the frequency of the $\sin$ wave encoding in the $\mathbf{L}$ matrix. Finally, we also compare with CoordConv~\cite{liu2018intriguing} approach. 

\begin{table}[htb]
\centering
\begin{tabular}{lr}
\begin{minipage}{.6\linewidth}
\begin{tabular}{@{}p{0.7cm}p{0.7cm}p{0.7cm}p{0.7cm}|c|c@{}}
\toprule
\multicolumn{4}{c|}{Features} &  \multirow{2}{*}{Frequency} & \multirow{2}{*}{mIoU} \\ 
$\mathbf{A}$ & $\mathbf{B}$ & $\mathbf{C}$ & $\mathbf{D}$ &   &  \\ \hline 
&   &   &  &   1 & 51.07 \\
\checkmark   &   &   &  &   1 & 51.44 \\
&  \checkmark  &   &   &   1 & 52.26 \\
&   &  \checkmark  &   &   1 & 52.12 \\
&   &   &  \checkmark  &   1 & 50.04 \\ \hline
&  \checkmark  &   &   &   2 & 51.14 \\
&  \checkmark  &   &     & 0.5 & 50.32 \\ \hline
\multicolumn{5}{c|}{ $\mathbf{A}$ + CoordConv~\cite{liu2018intriguing}(width+ height)}   & 51.12  
\\ \bottomrule
\multicolumn{6}{c}{(a)} 
\end{tabular}
\end{minipage} &
\begin{minipage}{.3\linewidth}
\begin{tabular}{cc}
\toprule
$\beta$ & mIoU \\ \hline
0&51.73\\
1&50.25 \\
0.5&51.83 \\
learning & 52.26
\\ \bottomrule
\multicolumn{2}{c}{(b)} 
\end{tabular}
\end{minipage} 
\end{tabular}
\caption{Shows ablation studies and the impact of hyper-parameters on a validation set of the WiSe dataset. Table (a) shows the result of segmentation by adding location encoding to the features $\mathbf{A}$, $\mathbf{B}$, $\mathbf{C}$, and $\mathbf{D}$. Table (b) shows the effect of manual $\beta$ values and as a learnable parameter.\label{ablationStudy}}
\end{table}

Table~\ref{ablationStudy} shows the ablation studies conducted in the proposed network architecture by changing the various hyper-parameters. The Table~\ref{ablationStudy}(a) shows that the location encoding on the feature vector outperformed the traditional CoordConv~\cite{liu2018intriguing} approach. The experiment also shows that adding location encoding to the intermediate feature $\mathbf{B}$ and $\mathbf{C}$ is comparatively better than adding it to $\mathbf{A}$ and $\mathbf{D}$. The best frequency of $\sin$ wave in the $\mathbf{L}$ is found as 1. While Table~\ref{ablationStudy} (b) shows the learnable $\beta$ value outperforms the manually assigned $\beta$ value.

\subsection{Comparison with State-of-the-Art Techniques}

Comparison results of the proposed network with state-of-the-art techniques are presented in Table~\ref{result1}. From the table, we observe that none of the attention mechanisms (e.g., {\sc han}et~\cite{Choi_2020_CVPR}, {\sc dan}et~\cite{fu2019dual}, and {\sc dran}et~\cite{fu2020scene}) is better than Haurilet \emph{et al.}~\cite{haurilet2019spase} on SPaSe dataset. While in the case of WiSe dataset, all attention architectures {\sc han}et~\cite{Choi_2020_CVPR}, {\sc dan}et~\cite{fu2019dual}, and {\sc dran}et~\cite{fu2020scene} are better than Haurilet \emph{et al.}~\cite{haurilet2019wise}. {\sc dran}et~\cite{fu2020scene} obtains the best performance ($35.77$\% mean IoU and $78.64$\% PA) among all existing techniques on SPaSe dataset. While {\sc dan}et~\cite{fu2019dual} obtains the best performance ($44.85$\% mean IoU and $88.80$\% PA) among all existing techniques on WiSe dataset. The proposed network obtains the best results ($46.74$\% mean IoU and $89.80$\% PA) on Wise and ($36.17$\% mean IoU and $79.11$\% PA) on SPaSe dataset compared to the existing techniques. 

\begin{table}[ht!]
\begin{center}
\begin{tabular}{||l||c|c||c|c||} \hline
\textbf{Methods} &\multicolumn{4}{|c||}{\textbf{Datasets}} \\ \cline{2-5}
&\multicolumn{2}{|c||}{\textbf{SPaSe}} &\multicolumn{2}{|c||}{\textbf{WiSe}} \\ \cline{2-5} 
&\textbf{mean IoU} &\textbf{PA}&\textbf{mean IoU} &\textbf{PA} \\ \hline \hline
Haurilet \emph{et al.}~\cite{haurilet2019spase} &35.80  &77.40 &- &- \\               
Haurilet \emph{et al.}~\cite{haurilet2019wise}  &-  &-  &37.20 &88.50 \\
HANet~\cite{Choi_2020_CVPR}                    &32.37  &76.54  &39.35 &88.72 \\
DANet~\cite{fu2019dual}                        &33.12  &77.39  &44.85 &88.80 \\ 
RANet~\cite{fu2020scene}                      &35.77  &78.64  &43.31 &88.77 \\
Ours                                          &36.17   &79.11         &46.74 &89.80 \\ \hline \hline
\end{tabular}
\end{center}
\caption{illustrates the comparison of the proposed method with state-of-the-art techniques on benchmark datasets WiSe and SPaSe datasets. \label{result1}}
\end{table}

\begin{figure}
\centering
\begin{subfigure}[b]{0.45\textwidth}
\centering
\includegraphics[width=\linewidth, height = 0.4\linewidth]{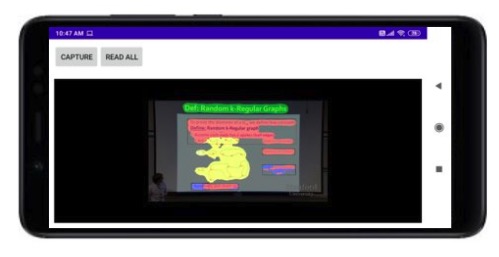}
\subcaption{}
\end{subfigure}
\hfill
\begin{subfigure}[b]{0.5\textwidth}
\centering
\includegraphics[width=\textwidth, trim={0cm  
0cm 
0cm 
0cm 
},clip]{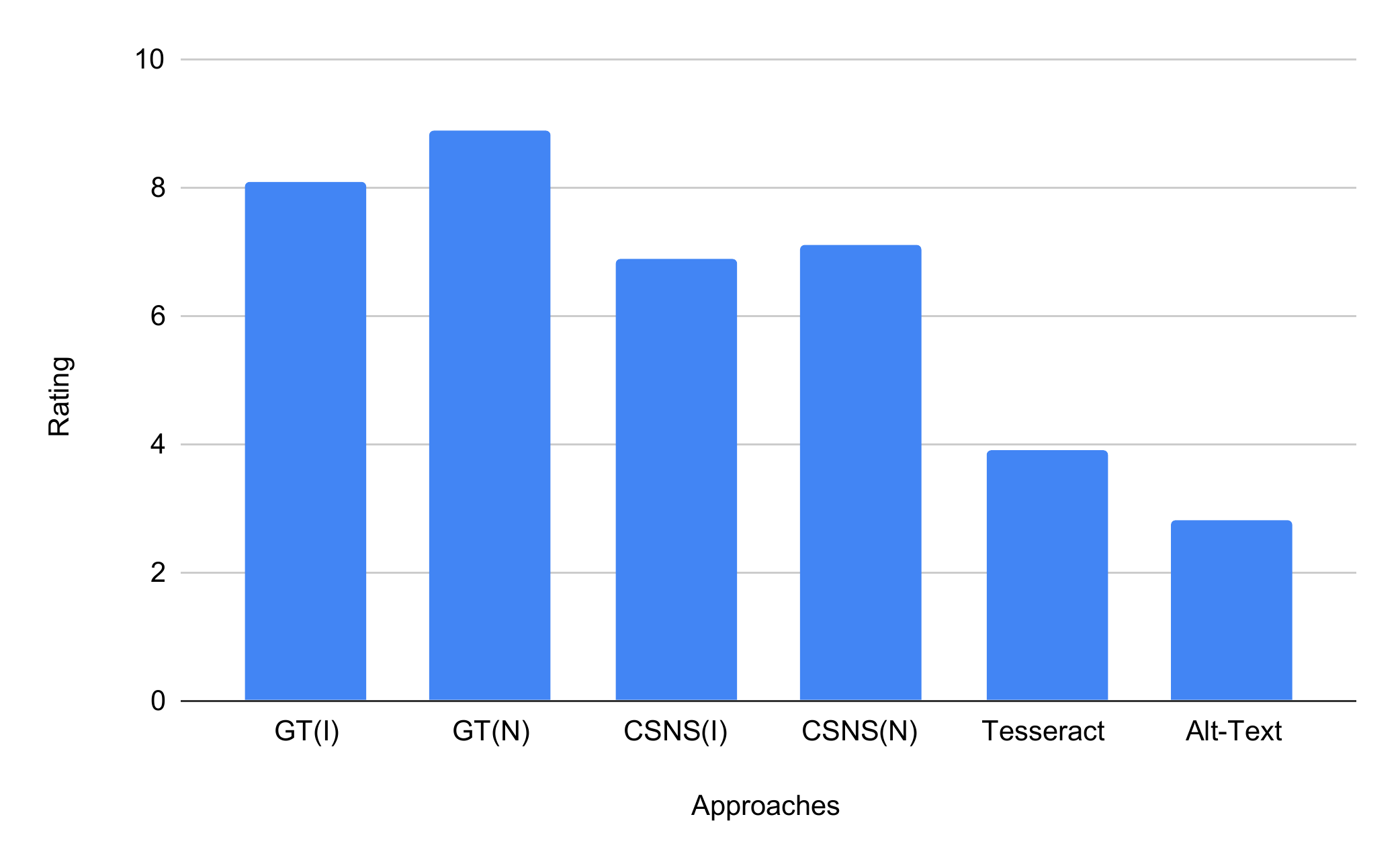}
\subcaption{}
\end{subfigure}
\caption{\textbf{(a):} presents user interface of the developed mobile application. The mobile screen displays the camera-captured slide image. Various colors highlight the slide's logical regions for better visualization. The {\sc Read All} button plays generated audio sound corresponding to the slide's content. \textbf{(b):} presents user rating on performance of various approaches. \textbf{I} and \textbf{N} indicate \textbf{interactive} and \textbf{non-interactive} modes, respectively.\label{fig_mobile_application}}
\end{figure}
\vspace{-0.08\textwidth}
\subsection{Narration System Evaluation}

We establish the effectiveness of the developed slide narration system by comparing the performance with Facebook’s Automatic Alt-Text (\textsc{aat})~\cite{wu2017automatic} and Tesseract \textsc{ocr}~\cite{smith2007overview}. We use $300$ test images of the WiSe dataset for evaluating the system. $30$ volunteers who can read and write English are selected. Each volunteer takes $300$ slide images and evaluates corresponding audio outputs generated by various techniques. The volunteer gives a grade (value in the range $[0,\ 10]$) to the generated audio files by various techniques by comparing them with the original slide content. We collect all these grades and plot them as a bar chart (shown in Fig.~\ref{fig_mobile_application}(b)). The user gives the best grade for the audio file corresponding to the ground truth segmentation of the slide. However, the proposed {\sc csns} also performs comparatively better than the existing techniques. 

\section{Summary}

This paper presents a classroom slide segmentation network to segment classroom slide images more accurately. The attention module in the proposed network enriches the learned contextual features by utilizing the location encoding of the logical region. We develop a Classroom Slide Narration System as an application of classroom slide segmentation for helping {\sc vi} students to read and understand the slide's content. Experiments on benchmark datasets WiSe and SpaSe conclude that the proposed {\sc lean}et achieves the best slide segmentation results as compared to the state-of-the-art methods. We also demonstrate the effectiveness of our developed Classroom Slide Narration System over existing assistive tools in terms of user experiences. 

\bibliographystyle{splncs04}
\bibliography{reference}

\end{document}